\documentclass[sigconf,balance=false,pbalance=true]{acmart}
\usepackage{graphicx}
\usepackage{amsmath}
\usepackage{algorithm}
\usepackage{multirow}
\usepackage{booktabs}
\usepackage{xcolor}
\usepackage{hyperref}
\newcommand{\best}[1]{\textcolor{red}{#1}}
\newcommand{\second}[1]{\textcolor{blue}{#1}}
\makeatletter
\newcommand{\authornotetext}[2]{\g@addto@macro\@authornotes{\footnotetext[#1]{#2}}}
\makeatother

\AtBeginDocument{%
  }
\setcopyright{cc}
\setcctype{by-nc-nd}
\copyrightyear{2026}
\acmYear{2026}
\acmDOI{10.1145/3767308.3836433}
\acmConference[MM '26]
  {Proceedings of the 34th ACM International Conference on Multimedia}
  {November 10--14, 2026}
  {Rio de Janeiro, Brazil}
\acmBooktitle{Proceedings of the 34th ACM International Conference on
  Multimedia (MM '26), November 10--14, 2026, Rio de Janeiro, Brazil}
\acmISBN{979-8-4007-2213-4/2026/11}

\acmSubmissionID{mfp8531}

\begin{document}

\title{TRACE: Training-time Report-guided and Clinically Ordered Concept Editing}
% for Robust Breast Ultrasound Diagnosis under Incomplete Concepts}

\author{Wentao Yue}
\authornotemark[2]
\affiliation{%
  \institution{Lanzhou University}
  \city{Lanzhou}
  \country{China}}
\email{yuewt21@lzu.edu.cn}

\author{Tianyou Lai}
\authornotemark[2]
\affiliation{%
  \institution{Lanzhou University}
  \city{Lanzhou}
  \country{China}}
\email{laity21@lzu.edu.cn}

\author{Jiayu Luo}
\affiliation{%
  \institution{Lanzhou University}
  \city{Lanzhou}
  \country{China}}
\email{luojy21@lzu.edu.cn}

\author{Qingyu Mao}
\affiliation{%
  \institution{Shenzhen University}
  \city{Shenzhen}
  \country{China}}
\email{2150432008@email.szu.edu.cn}

\author{Ziying Wang}
\affiliation{%
  \institution{Southern Medical University}
  \city{Guangzhou}
  \country{China}}
\email{ziyingwang@smu.edu.cn}

\author{Zhenyuan Ning}
\authornotemark[1]
\affiliation{%
  \institution{Southern Medical University}
  \city{Guangzhou}
  \country{China}}
\email{zhenyuan123@i.smu.edu.cn}

\author{Qilei Li}
\authornotemark[1]
\affiliation{%
  \institution{Central China Normal University}
  \city{Wuhan}
  \country{China}}
\email{qilei.li@ccnu.edu.cn}
\authornotetext{1}{Corresponding authors.}
\authornotetext{2}{These authors contributed equally to this work.}
\renewcommand{\shortauthors}{Wentao Yue et al.}

\begin{abstract}
Breast ultrasound diagnosis relies on clinically meaningful semantic concepts, yet most deep learning methods adopt end-to-end image-to-label paradigms that lack interpretability and robustness. While concept-based approaches offer a promising alternative, they often assume complete annotations or require multimodal inputs at inference, which significantly limits their real-world applicability. To tackle these issues, we propose Training-time Report-guided and Clinically Ordered Concept Editing (TRACE), a training-time report-guided framework that leverages structured radiology reports as privileged concept supervision while enabling image-only diagnosis at test time. TRACE refines image-derived concepts through a teacher-guided editing mechanism within a malignancy-aware ordered concept space. To address incomplete annotations, we introduce Strategic Concept Missing Training (SCMT) and train an image-only self-editor via edit distillation for autonomous concept refinement. Besides, we introduce BUSC, a concept-enriched benchmark linking images, labels, and structured attributes. Experiments across multiple datasets demonstrate that TRACE achieves superior performance and improved cross-domain robustness compared to existing methods. Code is available in our GitHub repository: 
\url{https://github.com/wentao-2/TRACE}

% \url{https://anonymous.4open.science/r/TRACE-1BE2}
\end{abstract}

%% ACM Computing Classification System metadata for this work.




\maketitle

\section{Introduction}

\begin{quote}
\large\itshape
``The eye sees only what the mind is prepared to comprehend.''
--- Robertson Davies
\end{quote}

In breast ultrasound diagnosis, reliable visual interpretation requires more than the detection of image cues. It also depends on clinically grounded concepts that organize visual evidence into diagnostic reasoning \cite{eghtedari2021current}. In clinical practice, radiologists do not make diagnoses solely from raw appearance patterns. Instead, they interpret lesions through structured semantic attributes such as shape, margin, orientation, posterior acoustic features, echogenicity, and calcification \cite{seely2026bi}. These concepts provide an intermediate abstraction between low-level image signals and final diagnostic decisions on benign or malignant status, which makes them highly valuable for robust and explainable breast ultrasound analysis \cite{yan2023robust}.

Despite substantial progress in deep neural networks for breast ultrasound diagnosis, most existing methods still follow an end-to-end image-to-label paradigm. Strong image-only backbones such as ResNet50 \cite{he2016deep} and ViT-B/16 \cite{dosovitskiy2020image} already achieve competitive performance, suggesting that current limitations may stem less from backbone capacity than from the lack of clinically grounded intermediate supervision \cite{xu2025samask}. Although such pipelines perform well in-domain, they often rely on shortcut visual correlations rather than clinically structured reasoning, limiting interpretability and robustness under distribution shifts or incomplete semantic supervision \cite{10530449}. Frequency and decomposed-spatial modeling has also been explored for industrial defect recognition \cite{zhou2026frednet}, but such representation-level approaches do not encode clinically grounded semantics. Concept-based learning therefore offers a more clinically aligned alternative by introducing semantically meaningful intermediate representations, as exemplified by VLG-CBM \cite{srivastava2024vlg}. However, existing frameworks typically assume either complete concept supervision or multimodal inference with text available at test time. In practice, structured reports or concept annotations may be accessible during model development, yet are often unavailable, incomplete, or inconsistently documented in real deployment.
\begin{figure}[t]
    \centering
    \includegraphics[width=1.05\linewidth]{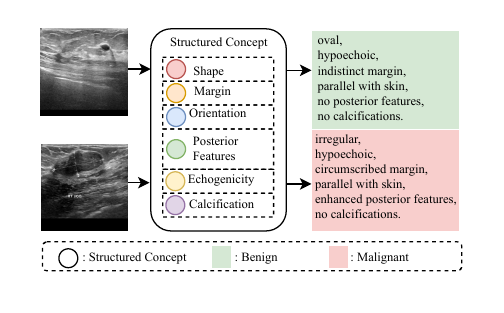}
    \includegraphics[width=0.9\linewidth]{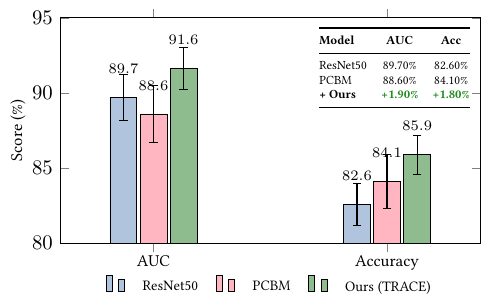}
    \caption{Overview of the structured concept representation used in BUSC-BUSBRA and performance comparison on the BUSC-BUSBRA dataset.}
    \Description{The upper panel shows structured breast ultrasound concepts, including shape, margin, orientation, posterior features, echogenicity, and calcification, together with benign and malignant examples. The lower panel compares ResNet50, PCBM, and TRACE on AUC and accuracy. TRACE achieves an AUC of 91.6 percent and an accuracy of 85.9 percent, improving over PCBM by 1.9 and 1.8 percentage points, respectively.}
    \label{fig:teaser}
\end{figure}
% ==================== 图表代码结束 ====================

A related challenge is the lack of standardized breast ultrasound resources that align images, diagnostic labels, and structured clinical concepts. Most public datasets support image-level diagnosis or lesion segmentation, while concept annotations are often sparse, inconsistent, or unavailable, limiting systematic study of concept-guided diagnosis under incomplete concepts \cite{ardakani2023open}. To address this gap, we construct the \textit{Breast Ultrasound Structured Concept (BUSC)} dataset, a concept-enriched benchmark that associates breast ultrasound images with structured semantic attributes and diagnostic labels. BUSC includes two development subsets: \textit{BUSC-BUSI647}, derived from the public BUSI \cite{al2020dataset} dataset by removing normal cases and retaining 647 lesion samples, and \textit{BUSC-BUSBRA}, built upon the public BUSBRA  \cite{gomez2024bus} dataset. These subsets serve as the primary benchmarks for model development and in-domain evaluation.

These observations raise a practical question: 
\textit{how can structured radiological reports be used during training to improve concept-level reasoning while supporting pure image-only diagnosis at test time?}
A straightforward solution is to treat reports as an additional text modality and fuse them with image features \cite{huang2021gloria,yuan2025reading}. However, this formulation does not fully capture their clinical function. In breast ultrasound, report fields encode curated semantic judgments that can help correct imperfect image-derived concepts. We therefore treat structured reports as \textit{privileged concept supervision} available only during training, rather than as an inference-time modality.

Inspired by these observations, we propose \textbf{TRACE}, a training-time report-guided and clinically ordered concept editing framework for robust breast ultrasound diagnosis under incomplete concepts. TRACE follows an asymmetric training--test paradigm: it uses breast ultrasound images together with report-derived concept annotations during training, but relies on images only at inference. Specifically, an image encoder first produces coarse concept predictions, after which a privileged concept teacher guides a teacher-driven editor to revise them. To better reflect clinical reasoning, TRACE organizes semantic attributes in a malignancy-aware ordered concept space, modeling diagnosis as clinically meaningful concept editing rather than flat independent classification.

To further handle incomplete concepts, TRACE introduces \textit{Strategic Concept Missing Training (SCMT)}, which masks report concepts during training according to clinically informed missing patterns. This encourages compensatory reasoning across correlated attributes instead of over-reliance on any single concept. Since reports are unavailable at test time, TRACE further trains an image-only self-editor via edit distillation to mimic the privileged teacher editor, enabling concept refinement and fully image-only diagnosis.

We evaluate TRACE on the two BUSC development subsets, \textit{BUSC-BUSBRA} and \textit{BUSC-BUSI647}, and further assess zero-shot cross-domain generalization on three external datasets: \textit{Ardakani}, \textit{BUS\_UC}, and \textit{BrEaST}. Compared with standard visual backbones, concept bottleneck models, prototype-based methods, explainable medical classifiers, and vision-language models, TRACE consistently achieves superior in-domain performance and strong cross-domain robustness. These results show that learning to revise clinically meaningful concepts from training-time report supervision is more effective than relying on either end-to-end visual prediction or inference-time multimodal assistance.

In summary, the main contributions of this work are as follows:
\begin{itemize}
\item We construct the \textbf{Breast Ultrasound Structured Concept (BUSC) }dataset, a concept-enriched benchmark for breast ultrasound that explicitly links images, diagnostic labels, and structured clinical concepts. BUSC contains two development subsets, \textit{BUSC-BUSI647} and \textit{-BUSBRA}, and provides a practical basis for the study of concept-guided diagnosis and robustness under incomplete concept supervision.
\item We propose \textbf{TRACE}, a training-time report-guided framework that treats structured radiology reports as a privileged concept teacher, thereby enabling stronger image-only diagnosis of breast ultrasound while requiring no reports at inference time.
\item We introduce \textbf{clinically ordered concept editing}, which represents semantic attributes in a malignancy-aware ordered concept space and explicitly learns how report supervision refines coarse concepts derived from images.
\item We develop \textbf{Strategic Concept Missing Training (SCMT)} together with an image-only self-editor and edit distillation, which improves robustness to incomplete concepts and supports generalizable image-only deployment across datasets.
\end{itemize}

\section{Related Work}

\textbf{Structured Concepts in Breast Ultrasound.}
Breast ultrasound diagnosis is naturally grounded in structured clinical descriptions. BI-RADS-related attributes, such as shape, margin, orientation, posterior acoustic features, and echogenicity, are routinely used for lesion assessment and report writing \cite{spak2017bi}. Recent studies have incorporated such structured descriptions into deep learning models to improve interpretability and clinical readability. For instance, BI-RADS-Net \cite{zhang2021bi} jointly predicts diagnostic labels and BI-RADS attributes via multi-task learning, while BI-RADS-Net-V2 \cite{zhang2023bi} further integrates semantic and quantitative explanations for more interpretable computer-aided diagnosis. Other works also build unified pipelines for detection, description, and classification based on expert-annotated BI-RADS terminology, highlighting the clinical value of structured attribute supervision.

However, most existing methods assume structured attributes or report information are available during evaluation, or rely on complete concept annotations within the same domain \cite{zhang2023bi} \cite{huang2021gloria}. In practice, breast ultrasound datasets often provide only images and class labels, without standardized concept annotations. Recent multimodal approaches incorporate radiology reports through image--text fusion, but still require reports as inference-time inputs \cite{wu2023medklip}. In contrast, we consider a setting where structured reports are only partially available during training, while testing and external generalization rely solely on images.
\begin{figure*}[t]
  \centering
  \includegraphics[width=0.85\textwidth]{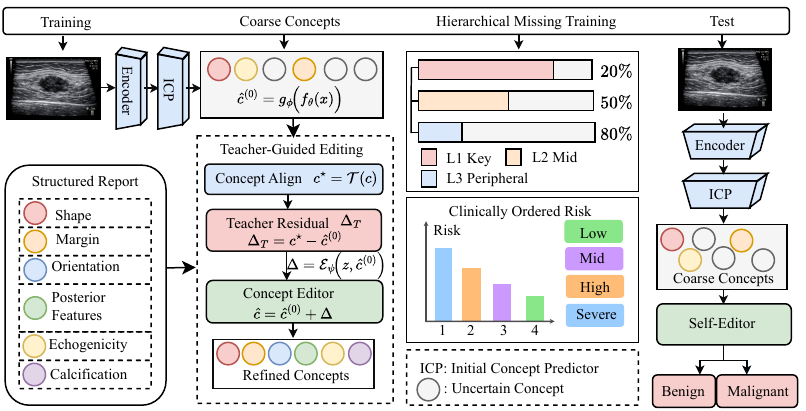}
  \caption{Overview of TRACE. During training, structured reports act as privileged concept teachers to correct coarse image-derived concepts through residual concept editing. TRACE further improves robustness with hierarchical concept missing training and a clinically ordered risk space. At test time, the report branch is removed, and diagnosis is made from images alone via self-editing concept refinement.}
  \Description{The TRACE framework consists of training and inference pathways. During training, breast ultrasound images produce coarse image-derived concepts, while structured reports provide privileged concept supervision for residual concept editing. Hierarchical concept missing training and a clinically ordered risk space improve robustness. During inference, the report branch is removed, and the image-only self-editor refines the concepts used for the final diagnosis.}

  \label{fig:main_framework}
\end{figure*}

\noindent
\textbf{Concept Bottleneck Models.}
Concept Bottleneck Models (CBMs) introduce a human-interpretable concept layer between perception and prediction, enabling concept-based explanations and human intervention \cite{sinha2025comprehensive}. The seminal work \cite{koh2020concept} of Koh et al. establishes the CBM paradigm and shows that such models maintain competitive performance while supporting concept-level interpretability.

Recent medical interpretable models further improve concept learning from different perspectives. Explicd \cite{gao2024aligning} learns interpretable concept representations by aligning images with textual diagnostic criteria provided by experts or large language models. VLG-CBM \cite{srivastava2024vlg} enhances concept faithfulness through vision-language guidance and grounded detectors. MVP-CBM \cite{wang2025mvp} improves concept bottleneck representations by modeling the preference of different concepts for features from different visual layers. Overall, these methods mainly focus on improving concept acquisition, concept--feature correspondence, or concept representation quality. In contrast, our goal is not simply to build a stronger CBM, but to address a practical breast ultrasound setting where structured reports are available only during training as privileged concept teachers, while diagnosis at test time relies on images alone.

\noindent
\textbf{Privileged Supervision.}
From a broader machine learning perspective, our problem is related to learning with additional information available during training but unavailable at test time. The Learning Using Privileged Information (LUPI) framework \cite{vapnik2015learning} proposed by Vapnik et al. shows that models can exploit extra teacher information during training while relying only on standard inputs at inference time. Subsequent work on generalized distillation further unifies knowledge distillation and privileged-information learning within a common framework.

This idea has also been explored in medical imaging \cite{chen2025medical}. For example, Yang et al. improve cross-modality image registration by introducing a third modality visible only during training \cite{yang2022cross}, and Proto-Caps \cite{gallee2025proto} combines prototype learning with privileged information for medical image classification. However, these methods typically treat extra information as an auxiliary source for distillation, a prototype-level constraint, or an extra modality. In contrast, we explicitly model structured breast ultrasound reports as concept-level teacher signals, aiming to learn a concept correction and diagnosis mechanism that transfers to image-only deployment.

\section{Methodology}

\subsection{Problem Setting and Method Overview}

We study an asymmetric clinical setting for breast ultrasound classification. Given a breast ultrasound image $x \in \mathcal{X}$, its class label is denoted by $y \in \mathcal{Y}$, where $\mathcal{Y}=\{0,1\}$ represents benign and malignant cases, respectively. For a subset of training samples, an expert-annotated structured concept vector is also available,
\begin{equation}
c=[c^{(1)},c^{(2)},\dots,c^{(M)}] \in \mathcal{C},
\end{equation}

where $M$ denotes the number of concepts. Each concept dimension corresponds to a clinically meaningful BI-RADS-related attribute in breast ultrasound, such as shape, margin, orientation and calcification. Unlike multimodal methods that treat structured text as an inference-time input modality, we model it as a privileged concept teacher accessible only during training, while test-time diagnosis and external generalization rely on images alone.

Formally, the training set consists of two parts:
\begin{equation}
\mathcal{D}_{s}=\{(x_i,c_i,y_i)\}_{i=1}^{N_s}, \qquad
\mathcal{D}_{w}=\{(x_j,y_j)\}_{j=1}^{N_w},
\end{equation}
where $\mathcal{D}_{s}$ denotes the sample set with structured concept supervision, and $\mathcal{D}_{w}$ denotes the weakly supervised sample set with only images and class labels. The test set is denoted by
\begin{equation}
\mathcal{D}_{\mathrm{test}}=\{(x_k,y_k)\}_{k=1}^{N_t},
\end{equation}
where no structured concepts or report information are provided. Our goal is to learn a mapping
\begin{equation}
f:\mathcal{X}\rightarrow \mathcal{Y},
\end{equation}
which uses the limited structured concept supervision in $\mathcal{D}_{s}$ during training while still enabling robust classification and concept reasoning from images alone at deployment.

The core idea of TRACE is to learn concept correction rather than concept prediction alone. Specifically, the model extracts visual representations and produces initial image-derived concepts, then uses structured reports during training as concept-level teacher signals to learn a correction mapping from coarse to more reliable concepts. At test time, the learned correction behavior is applied using images only, enabling report-free deployment. The overall pipeline is formulated as
\begin{equation}
z=f_{\theta}(x), \qquad
\hat{c}^{(0)}=g_{\phi}(z), \qquad
\hat{c}=\mathcal{E}(z,\hat{c}^{(0)}), \qquad
\hat{y}=h_{\omega}(\hat{c}),
\end{equation}
where $f_{\theta}$ is the image encoder, $g_{\phi}$ is the initial concept predictor, $\mathcal{E}$ is the concept editor, and $h_{\omega}$ is the final classifier. During training, $\mathcal{E}$ learns concept correction under the guidance of the privileged concept teacher. At test time, $\mathcal{E}$ reduces to a self-editor that depends only on the image and the initial concepts.

\subsection{Report-Guided Concept Editing}
Given an input image $x$, the image encoder $f_{\theta}$ extracts a visual feature and the initial concept predictor $g_{\phi}$ produces a coarse concept representation:
\begin{equation}
\begin{aligned}
z&=f_{\theta}(x)\in\mathbb{R}^{d},\\
\hat{c}^{(0)}&=g_{\phi}(z)=\big[\hat{c}^{(0,1)},\hat{c}^{(0,2)},\dots,\hat{c}^{(0,M)}\big].
\end{aligned}
\end{equation}
For discrete concepts, $\hat{c}^{(0,m)}$ can be represented as class logits or probability distributions. For ordered concepts, $\hat{c}^{(0,m)}$ can also be represented as ordered real-valued scores or ordered embeddings. Because image appearance is affected by noise, device variation, and domain shift, concepts predicted directly by $g_{\phi}$ are often coarse and unstable, motivating subsequent concept-level correction.

For samples with structured annotations $(x,c,y)\in\mathcal{D}_s$, we do not treat $c$ as a test-time input modality. Instead, we regard it as a concept-level teacher signal available only during training. We first transform the structured concept annotation into the unified concept space and compute the residual correction target:
\begin{equation}
c^{T} = \mathcal{T}(c),\qquad
\Delta_{T} = c^{T}-\hat{c}^{(0)},
\end{equation}
where $\mathcal{T}(\cdot)$ maps the original structured concept annotation into the unified concept representation space. The teacher concept $c^{T}$ specifies the desired semantic state, while $\Delta_T$ represents the correction direction and magnitude required to refine the initial image-derived concepts. Therefore, TRACE learns to estimate concept corrections from structured clinical knowledge rather than directly predicting a new set of concepts from scratch.

We introduce a concept editor $\mathcal{E}_{\psi}$, which takes the visual feature $z$ and concept $\hat{c}^{(0)}$ as input and outputs a concept correction term:
\begin{equation}
\Delta_{\psi}=\mathcal{E}_{\psi}\big(z,\hat{c}^{(0)}\big),\qquad
\hat{c}=\hat{c}^{(0)}+\Delta_{\psi}.
\end{equation}
During training, for samples with structured concept supervision, we minimize the discrepancy between $\Delta_{\psi}$ and the teacher residual $\Delta_T$, so that the editor learns how to correct rather than merely where to end up. Accordingly, the editing constraint is defined as
\begin{equation}
\mathcal{L}_{\mathrm{edit}}
=
\frac{1}{M}\sum_{m=1}^{M}\ell_{\mathrm{edit}}
\Big(
\Delta_{\psi}^{(m)},\,
\Delta_T^{(m)}
\Big),
\end{equation}
where $\ell_{\mathrm{edit}}(\cdot,\cdot)$ denotes the concept-dependent editing loss.

Several concepts in breast ultrasound are not independent flat discrete labels, but follow clear clinical risk order. For example, concepts such as margin and shape correspond to different levels of malignancy risk. We therefore introduce clinically ordered constraints into the concept space so that the editing direction is more consistent with medical knowledge rather than arbitrary label switching. For the $m$-th ordered concept, let its ordinal level be
\begin{equation}
r^{(m)} \in \{1,2,\dots,K_m\},
\end{equation}
where $K_m$ denotes the number of ordinal levels for that concept, and let the predicted ordinal score be $s^{(m)}$. We then define the ordinal loss as
\begin{equation}
\mathcal{L}_{\mathrm{ord}}
=
\sum_{m\in\mathcal{M}_{\mathrm{ord}}}
\ell_{\mathrm{ord}}\big(s^{(m)},r^{(m)}\big),
\end{equation}
where $\mathcal{M}_{\mathrm{ord}}$ denotes the set of all concepts with clinical risk order. This constraint encourages concept correction to follow clinically plausible directions and reduces label transitions that are inconsistent with medical knowledge.

The final edited concept representation is denoted as $\hat{c}$ and used for downstream diagnosis:
\begin{equation}
\hat{y}=h_{\omega}(\hat{c}).
\end{equation}
\subsection{Hierarchical Concept Missing Training}

Structured concepts in real clinical reports are unevenly available, and the missing probabilities vary substantially across attributes. Rather than simple random masking, we adopt a hierarchical concept missing training strategy to simulate supervision sparsity better reflecting real report patterns. Specifically, we divide all concepts into three levels by importance and clinical availability:
\begin{equation}
\mathcal{C}=\mathcal{C}_{L1}\cup \mathcal{C}_{L2}\cup \mathcal{C}_{L3},
\end{equation}
where $\mathcal{C}_{L1}$, $\mathcal{C}_{L2}$, and $\mathcal{C}_{L3}$ denote the key, mid-level, and peripheral concept sets, respectively. In our setting, the hierarchy is
\begin{equation}
\begin{aligned}
\mathcal{C}_{L1} &= \{\mathrm{margin},\mathrm{shape}\},\\
\mathcal{C}_{L2} &= \{\mathrm{orientation},\mathrm{posterior},\mathrm{echogenicity}\},\\
\mathcal{C}_{L3} &= \{\mathrm{calcification}\}.
\end{aligned}
\end{equation}

For concepts at level $L$, we define the missing rate as
\begin{equation}
\begin{aligned}
r_L &= 0.2 + 0.3(L - 1), \qquad L \in \{1,2,3\},\\
r_1 &= 0.2,\qquad r_2 = 0.5,\qquad r_3 = 0.8.
\end{aligned}
\end{equation}
Thus, key concepts are more likely to be preserved, while peripheral concepts are more likely to be missing. For concept $c^{(m)}$, we define a visibility mask
\begin{equation}
b^{(m)}\sim \mathrm{Bernoulli}\big(1-r_{L_m}\big),
\end{equation}
where $L_m$ denotes the level of the $m$-th concept. The masked teacher concept is
\begin{equation}
\tilde{c}^{(m)} = b^{(m)} c^{(m)} + \big(1-b^{(m)}\big)\varnothing,
\end{equation}
where $\varnothing$ denotes a missing concept. With the full mask vector
\begin{equation}
b=[b^{(1)},b^{(2)},\dots,b^{(M)}], \qquad
\tilde{c}=b\odot c + (1-b)\odot \varnothing.
\end{equation}

During training, the editor accesses only unmasked teacher signals, so the editing target becomes
\begin{equation}
\Delta_T^{\mathrm{mask}} = b\odot \Delta_T.
\end{equation}
The mask-aware editing loss is
\begin{equation}
\mathcal{L}_{\mathrm{mask\_edit}}
=
\frac{1}{\sum_{m=1}^{M} b^{(m)}+\epsilon}
\sum_{m=1}^{M}
b^{(m)}\,
\ell_{\mathrm{edit}}
\Big(
\Delta_{\psi}^{(m)},\,
\Delta_T^{(m)}
\Big),
\end{equation}
where $\epsilon$ is a small constant to prevent division by zero. This design forces the model to learn cross-concept dependency and redundancy compensation under incomplete concept supervision, reducing the gap between training with visible structured reports and testing without reports.
% We note that a stronger form of supervision scarcity arises when only a subset of training samples has any structured report annotations, that is, when the supervision ratio is below $100\%$. This setting corresponds to complete report absence for some samples rather than partial concept-level missingness. Since it is more suitable as a robustness study of annotation scarcity, we investigate it separately in the ablation experiments instead of treating it as a core training mechanism of TRACE.

\subsection{Inference and Training Objectives}
At test time, structured reports are unavailable and the teacher branch is removed. The model performs self-editing using only image features and initial concepts. Given a test image $x$, we compute
\begin{equation}
z=f_{\theta}(x),\qquad
\hat{c}^{(0)}=g_{\phi}(z),\qquad
\Delta_{S}=\mathcal{E}_{\psi}\big(z,\hat{c}^{(0)}\big),
\end{equation}
which gives final concept representation and classification output
\begin{equation}
\hat{c}=\hat{c}^{(0)}+\Delta_S,\qquad
\hat{y}=h_{\omega}(\hat{c}).
\end{equation}
Thus, the goal of TRACE is not to recover structured reports at test time, but to transfer the correction behavior learned from the privileged concept teacher during training into report-free concept refinement.

For the main classification task, we use the supervised loss
\begin{equation}
\mathcal{L}_{\mathrm{cls}}
=
\ell_{\mathrm{cls}}(\hat{y},y),
\end{equation}
where $\ell_{\mathrm{cls}}$ is typically the binary cross-entropy loss.

For training samples with concept annotations, the initial concept predictor is also trained to fit the teacher concepts. We define the initial concept supervision loss as
\begin{equation}
\mathcal{L}_{\mathrm{init}}
=
\frac{1}{M}\sum_{m=1}^{M}
\ell_{\mathrm{con}}
\Big(
\hat{c}^{(0,m)},\,
c^{\star(m)}
\Big).
\end{equation}
This term keeps the initial image-derived concepts close to the clinical concept space, providing a reasonable starting point for subsequent editing.

For samples with structured concept supervision, we use the previously defined $\mathcal{L}_{\mathrm{mask\_edit}}$ to supervise the editing residual. To reduce the mismatch between teacher-guided editing during training and self-editing at test time, we also introduce a post-editing consistency constraint
\begin{equation}
\mathcal{L}_{\mathrm{cons}}
=
\frac{1}{M}\sum_{m=1}^{M}
\ell_{\mathrm{con}}
\Big(
\hat{c}^{(m)},\,
\tilde{c}^{(m)}
\Big),
\end{equation}
where the loss is computed only on currently visible teacher concept dimensions. This term encourages the self-edited concept state to approach the target concept state induced by teacher supervision.

Combining all terms above, the training objective of TRACE is
\begin{equation}
\mathcal{L}
=
\mathcal{L}_{\mathrm{cls}}
+\lambda_{1}\mathcal{L}_{\mathrm{init}}
+\lambda_{2}\mathcal{L}_{\mathrm{mask\_edit}}
+\lambda_{3}\mathcal{L}_{\mathrm{ord}}
+\lambda_{4}\mathcal{L}_{\mathrm{cons}},
\end{equation}
where $\lambda_{1},\lambda_{2},\lambda_{3},\lambda_{4}\ge 0$ are balancing coefficients. For samples without concept supervision, namely $(x,y)\in\mathcal{D}_{w}$, the model optimizes only $\mathcal{L}_{\mathrm{cls}}$. For samples with concept supervision, namely $(x,c,y)\in\mathcal{D}_{s}$, it jointly optimizes the concept-related terms. In this way, TRACE learns a transferable concept correction mechanism from limited structured concept supervision and enables robust breast ultrasound diagnosis from images alone at test time.

\begin{table*}[t]
\centering
\scriptsize
\caption{Main comparison on BUSC-BUSBRA and BUSC-BUSI647. We compare TRACE with eight representative baselines. Best and second-best results in each column are highlighted in \textcolor{red}{red} and \textcolor{blue}{blue}, respectively.}
\label{tab:main_comparison}
\setlength{\tabcolsep}{4.5pt}
\renewcommand{\arraystretch}{1.03}
\resizebox{0.95\textwidth}{!}{%
\begin{tabular}{llcccccc}
\toprule
\multirow{2}{*}{\textbf{Setting}} & \multirow{2}{*}{\textbf{Model}} 
& \multicolumn{3}{c}{\textbf{BUSC-BUSBRA}} 
& \multicolumn{3}{c}{\textbf{BUSC-BUSI647}} \\
\cmidrule(lr){3-5} \cmidrule(lr){6-8}
& & \textbf{AUC} & \textbf{Acc} & \textbf{F1} 
& \textbf{AUC} & \textbf{Acc} & \textbf{F1} \\
\midrule

\multirow{2}{*}{\textbf{Black-box}}
& ResNet50 \cite{he2016deep}
& \second{0.897$\pm$0.015} & 0.826$\pm$0.014 & 0.725$\pm$0.027 
& 0.945$\pm$0.027 & 0.901$\pm$0.026 & 0.838$\pm$0.047 \\

& ViT-B/16 \cite{dosovitskiy2020image}
& 0.876$\pm$0.018 & 0.821$\pm$0.025 & 0.711$\pm$0.042 
& 0.944$\pm$0.021 & 0.886$\pm$0.035 & 0.809$\pm$0.068 \\
\midrule

\textbf{Prototype}
& ProtoCaps \cite{gallee2025proto}
& 0.727$\pm$0.037 & 0.718$\pm$0.024 & 0.391$\pm$0.075 
& 0.803$\pm$0.024 & 0.807$\pm$0.014 & 0.658$\pm$0.052 \\
\midrule

\multirow{3}{*}{\textbf{CBM}}
& PCBM \cite{yuksekgonul2022post}
& 0.886$\pm$0.019 & \second{0.841$\pm$0.018} & 0.738$\pm$0.031 
& \second{0.951$\pm$0.022} & 0.901$\pm$0.025 & 0.837$\pm$0.047 \\

& MVP-CBM \cite{wang2025mvp}
& 0.889$\pm$0.020 & 0.830$\pm$0.021 & \second{0.749$\pm$0.027} 
& 0.931$\pm$0.027 & 0.878$\pm$0.045 & 0.806$\pm$0.069 \\

& VLG-CBM \cite{srivastava2024vlg}
& 0.883$\pm$0.018 & 0.837$\pm$0.024 & 0.733$\pm$0.037 
& 0.941$\pm$0.028 & \second{0.910$\pm$0.018} & \second{0.857$\pm$0.029} \\
\midrule

\textbf{Explainable}
& Explicd \cite{gao2024aligning}
& 0.891$\pm$0.025 & 0.833$\pm$0.027 & 0.730$\pm$0.050 
& 0.939$\pm$0.029 & 0.884$\pm$0.016 & 0.822$\pm$0.034 \\
\midrule

\textbf{Foundation VLM}
& CLIP \cite{radford2021learning}
& 0.553$\pm$0.033 & 0.662$\pm$0.011 & 0.111$\pm$0.042 
& 0.536$\pm$0.020 & 0.674$\pm$0.010 & 0.018$\pm$0.036 \\
\midrule

\textbf{Ours}
& \textbf{TRACE}
& \best{0.916$\pm$0.014} & \best{0.859$\pm$0.013} & \best{0.776$\pm$0.026} 
& \best{0.970$\pm$0.030} & \best{0.938$\pm$0.057} & \best{0.902$\pm$0.083} \\
\bottomrule
\end{tabular}%
}
\end{table*}
\section{Experiments}

\subsection{Experimental Setup}

\subsubsection{Datasets}
We evaluate TRACE on five breast ultrasound datasets, including two subsets from our \textbf{Breast Ultrasound Structured Concept (BUSC)} dataset and three external datasets for cross-domain evaluation. Specifically, we augment the public BUSBRA \cite{gomez2024bus} and BUSI \cite{al2020dataset} datasets with expert-annotated structured semantic concepts to construct BUSC, which aligns each ultrasound image with diagnostic labels and structured concepts. \textbf{BUSC-BUSBRA}, built upon BUSBRA, contains 1,872 cases annotated with six semantic concepts: \textit{shape}, \textit{margin}, \textit{orientation}, \textit{posterior acoustic features}, \textit{echo pattern}, and \textit{calcification}. \textbf{BUSC-BUSI647}, derived from BUSI after removing normal samples, contains 647 lesion cases annotated with the same concept schema. These subsets are used for in-domain evaluation.

For external validation, we adopt three public breast ultrasound datasets from different acquisition sources and populations. \textbf{Ardakani} \cite{ardakani2023open} contains 232 pathologically confirmed lesions from Iran, including 123 malignant and 109 benign cases. \textbf{BUS\_UC} \cite{iqbal2024memory} contains 811 images collected in Pakistan, including 358 benign and 453 malignant cases. \textbf{BrEaST} \cite{pawlowska2024curated} is a curated dataset from Poland containing 256 scans with benign, malignant, and a small number of normal cases.

For cross-domain evaluation, we remove structured reports and concept annotations from \textbf{BUSC-BUSI647} and denote the resulting dataset as \textbf{BUSC-BUSI647$^{*}$}. Together with \textbf{Ardakani}, \textbf{BUS\_UC}, and \textbf{BrEaST}, it is treated as a target domain without concept supervision. This setup reflects realistic deployment, where structured reports and concept annotations are available only during training, while inference follows an image-only protocol.

\subsubsection{Compared Methods}
We compare TRACE with eight representative baselines. ResNet50 \cite{he2016deep} is used as a standard CNN baseline for image-based classification without concept supervision. ViT-B/16 \cite{dosovitskiy2020image} represents transformer-based image classification. ProtoCaps \cite{gallee2025proto} is included as a prototype-based medical image classifier. PCBM \cite{yuksekgonul2022post} serves as a standard concept bottleneck baseline, while MVP-CBM \cite{wang2025mvp} represents a stronger CBM variant with multi-view prototype learning. VLG-CBM \cite{srivastava2024vlg} is adopted as a vision-language-guided concept bottleneck model, and Explicd \cite{gao2024aligning} is included as an explainable medical image classifier. CLIP \cite{radford2021learning} serves as a representative foundation vision-language model based on contrastive pretraining.

\subsubsection{Implementation Details}
To ensure fair comparison, all BUSC images are processed with \textbf{zero-padding}, where black borders are added at the bottom to standardize spatial resolution while preserving the original aspect ratio. Unless otherwise specified, the image encoder is initialized with ImageNet-pretrained weights. For in-domain experiments, we adopt five-fold cross-validation and report the mean and standard deviation over the five folds. During training, TRACE uses ultrasound images with concept supervision derived from structured reports, while inference uses images only. The concept editor is optimized jointly with the image encoder and diagnosis head, and SCMT is applied during training to simulate incomplete concept conditions.

\subsubsection{Evaluation Metrics}
We report standard classification metrics for breast ultrasound diagnosis, including AUC, Accuracy, and F1-score. For in-domain evaluation, all methods are compared on BUSC-BUSBRA and -BUSI647 under the unified image-only inference setting. For external evaluation, models trained on the BUSC subsets are directly transferred to Ardakani, BUS\_UC, BrEaST, and BUSI647$^*$ to assess zero-shot cross-domain generalization.

\label{sec:setup}

\begin{table*}[t]
\centering
\caption{5-fold cross-domain zero-shot performance (\textit{mean}$\pm$\textit{std}). Best and second-best results are highlighted in \textcolor{red}{red} and \textcolor{blue}{blue}.}
\label{tab:complete_results}
\setlength{\tabcolsep}{7pt}
\renewcommand{\arraystretch}{1.15}

\resizebox{\textwidth}{!}{%
\begin{tabular}{lcccccccc}
\toprule
\multirow{2}{*}{\textbf{Model}} 
& \multicolumn{2}{c}{\textbf{Ardakani}} 
& \multicolumn{2}{c}{\textbf{BUS\_UC}} 
& \multicolumn{2}{c}{\textbf{BrEaST}} 
& \multicolumn{2}{c}{\textbf{BUSC-BUSI647$^{*}$}} \\
\cmidrule(lr){2-3} \cmidrule(lr){4-5} \cmidrule(lr){6-7} \cmidrule(lr){8-9}
& \textbf{AUC} & \textbf{Acc} & \textbf{AUC} & \textbf{Acc} 
& \textbf{AUC} & \textbf{Acc} & \textbf{AUC} & \textbf{Acc} \\
\midrule

ResNet50 \cite{he2016deep}
& 0.857$\pm$0.013 & 0.794$\pm$0.025 
& 0.525$\pm$0.074 & 0.470$\pm$0.053 
& 0.508$\pm$0.075 & 0.515$\pm$0.125 
& \second{0.873$\pm$0.007} & 0.819$\pm$0.016 \\

ViT-B16 \cite{dosovitskiy2020image}
& 0.844$\pm$0.016 & \second{0.825$\pm$0.017} 
& \second{0.657$\pm$0.030} & \best{0.622$\pm$0.026} 
& 0.771$\pm$0.019 & \second{0.725$\pm$0.029} 
& 0.847$\pm$0.015 & 0.813$\pm$0.012 \\

ProtoCaps \cite{gallee2025proto}
& 0.756$\pm$0.015 & 0.792$\pm$0.016 
& 0.538$\pm$0.025 & 0.518$\pm$0.024 
& 0.671$\pm$0.023 & 0.655$\pm$0.008 
& 0.764$\pm$0.015 & 0.764$\pm$0.010 \\

PCBM \cite{yuksekgonul2022post}
& 0.851$\pm$0.015 & 0.808$\pm$0.051 
& 0.653$\pm$0.042 & 0.589$\pm$0.051 
& 0.779$\pm$0.111 & 0.691$\pm$0.096 
& 0.864$\pm$0.010 & \second{0.826$\pm$0.015} \\

MVP-CBM \cite{wang2025mvp}
& \best{0.863$\pm$0.008} & 0.776$\pm$0.027 
& 0.647$\pm$0.032 & 0.591$\pm$0.036 
& 0.769$\pm$0.104 & 0.664$\pm$0.118 
& 0.871$\pm$0.007 & 0.806$\pm$0.016 \\

VLG-CBM \cite{srivastava2024vlg}
& 0.845$\pm$0.016 & 0.800$\pm$0.040 
& 0.651$\pm$0.012 & 0.593$\pm$0.016 
& 0.779$\pm$0.058 & 0.711$\pm$0.071 
& 0.867$\pm$0.010 & 0.825$\pm$0.010 \\

Explicd \cite{gao2024aligning}
& 0.850$\pm$0.012 & 0.793$\pm$0.020 
& 0.635$\pm$0.044 & 0.587$\pm$0.037 
& \second{0.785$\pm$0.100} & 0.715$\pm$0.121 
& 0.872$\pm$0.007 & 0.815$\pm$0.009 \\

CLIP \cite{radford2021learning}
& 0.621$\pm$0.032 & 0.733$\pm$0.013 
& 0.472$\pm$0.021 & 0.445$\pm$0.055 
& 0.661$\pm$0.094 & 0.621$\pm$0.046 
& 0.5357$\pm$0.020 & 0.674$\pm$0.010 \\

\midrule

\textbf{TRACE (Ours)}
& \second{0.862$\pm$0.018} & \best{0.836$\pm$0.046} 
& \best{0.661$\pm$0.027} & \second{0.604$\pm$0.023} 
& \best{0.822$\pm$0.019} & \best{0.757$\pm$0.023} 
& \best{0.873$\pm$0.023} & \best{0.857$\pm$0.017} \\

\bottomrule
\end{tabular}
}
\end{table*}

\begin{table}[t]
\centering
\scriptsize
\caption{Cross-domain comparison of different TRACE editors. Models are trained on BUSBRA and tested on external datasets without target-domain fine-tuning. Best and second-best results are highlighted in \textcolor{red}{red} and \textcolor{blue}{blue}.}
\label{tab:ablation_cross_editor}
\setlength{\tabcolsep}{3.5pt}
\renewcommand{\arraystretch}{1.03}
\resizebox{\columnwidth}{!}{%
\begin{tabular}{llcc}
\toprule
\textbf{Target} & \textbf{Editor} & \textbf{AUC} & \textbf{Acc} \\
\midrule
\multirow{3}{*}{\textbf{Ardakani}}
& TRACE-MLP  & \best{0.862$\pm$0.011} & 0.833$\pm$0.026 \\
& TRACE-Att  & \best{0.862$\pm$0.018} & \second{0.836$\pm$0.046} \\
& TRACE-Gate & \second{0.861$\pm$0.014} & \best{0.853$\pm$0.017} \\
\midrule
\multirow{3}{*}{\textbf{BUS\_UC}}
& TRACE-MLP  & \second{0.660$\pm$0.011} & \best{0.614$\pm$0.008} \\
& TRACE-Att  & \best{0.661$\pm$0.027} & \second{0.604$\pm$0.023} \\
& TRACE-Gate & 0.645$\pm$0.031 & 0.589$\pm$0.018 \\
\midrule
\multirow{3}{*}{\textbf{BrEaST}}
& TRACE-MLP  & 0.813$\pm$0.015 & 0.741$\pm$0.031 \\
& TRACE-Att  & \best{0.822$\pm$0.019} & \best{0.757$\pm$0.023} \\
& TRACE-Gate & \second{0.819$\pm$0.021} & \second{0.748$\pm$0.021} \\
\midrule
\multirow{3}{*}{\textbf{BUSI647$^{*}$}}
& TRACE-MLP  & \best{0.8736$\pm$0.017} & 0.8496$\pm$0.029 \\
& TRACE-Att  & \second{0.8733$\pm$0.023} & \second{0.8574$\pm$0.017} \\
& TRACE-Gate & 0.8711$\pm$0.018 & \best{0.8651$\pm$0.030} \\
\bottomrule
\end{tabular}%
}
\end{table}

\subsection{Main Results}
\label{sec:mainres}

Table~\ref{tab:main_comparison} summarizes the comparison between TRACE and eight representative baselines on BUSC-BUSBRA and -BUSI647. All methods are evaluated under a unified \textbf{image-only} test protocol, where structured reports and concept labels are used only during training. TRACE achieves the best results on both datasets, demonstrating the effectiveness of using structured reports as privileged concept teachers for image-only breast ultrasound diagnosis.

On BUSC-BUSBRA, TRACE achieves the best results on AUC, Acc, and F1, outperforming both image-only baselines such as ResNet50 and concept-based methods including PCBM, MVP-CBM, and Explicd. The advantage of TRACE is further evident on BUSC-BUSI647, where it again achieves the best results on all three metrics with clear improvements over competing methods. These results indicate that TRACE effectively exploits structured semantic supervision during training and transfers it into stronger image-only concept refinement at test time.

\begin{table*}[t]
\centering
\scriptsize
\caption{Core ablation of TRACE on BUSBRA and BUSC-BUSI647. We compare three editor designs under two training settings: \textit{Full} (without concept missing) and \textit{Missing} (with concept missing). Best and second-best results within each dataset and each setting are marked in \textcolor{red}{red} and \textcolor{blue}{blue}, respectively.}
\label{tab:ablation_combined}
\setlength{\tabcolsep}{3pt}
\renewcommand{\arraystretch}{1.02}
\resizebox{0.85\textwidth}{!}{%
\begin{tabular}{llcccccc}
\toprule
\multirow{2}{*}{\textbf{Setting}} & \multirow{2}{*}{\textbf{Editor}} 
& \multicolumn{3}{c}{\textbf{BUSC-BUSBRA}} 
& \multicolumn{3}{c}{\textbf{BUSC-BUSI647}} \\
\cmidrule(lr){3-5} \cmidrule(lr){6-8}
& & \textbf{AUC} & \textbf{Acc} & \textbf{F1} 
  & \textbf{AUC} & \textbf{Acc} & \textbf{F1} \\
\midrule

\multirow{3}{*}{\textbf{Full}}
& Attention 
& \best{0.924$\pm$0.007} & \best{0.874$\pm$0.016} & \best{0.794$\pm$0.023}
& 0.945$\pm$0.023 & 0.893$\pm$0.024 & 0.822$\pm$0.071 \\

& Gating    
& \second{0.921$\pm$0.014} & \second{0.859$\pm$0.012} & \second{0.783$\pm$0.012}
& \second{0.948$\pm$0.016} & \second{0.902$\pm$0.018} & \second{0.841$\pm$0.036} \\

& MLP       
& 0.916$\pm$0.014 & \second{0.859$\pm$0.013} & 0.776$\pm$0.026
& \best{0.970$\pm$0.030} & \best{0.938$\pm$0.057} & \best{0.902$\pm$0.083} \\

\midrule

\multirow{3}{*}{\textbf{Missing}}
& Attention 
& \second{0.924$\pm$0.013} & \best{0.874$\pm$0.019} & \best{0.802$\pm$0.018}
& \second{0.873$\pm$0.023} & \second{0.857$\pm$0.017} & \second{0.759$\pm$0.025} \\

& Gating    
& \best{0.926$\pm$0.012} & 0.866$\pm$0.017 & 0.784$\pm$0.031
& 0.871$\pm$0.018 & \best{0.865$\pm$0.030} & \best{0.763$\pm$0.042} \\

& MLP       
& 0.921$\pm$0.010 & \second{0.872$\pm$0.006} & \second{0.797$\pm$0.013}
& \best{0.874$\pm$0.017} & 0.850$\pm$0.029 & 0.745$\pm$0.036 \\

\bottomrule
\end{tabular}%
}
\end{table*}
\begin{figure*}[t]
  \centering
  \includegraphics[width=0.98\textwidth]{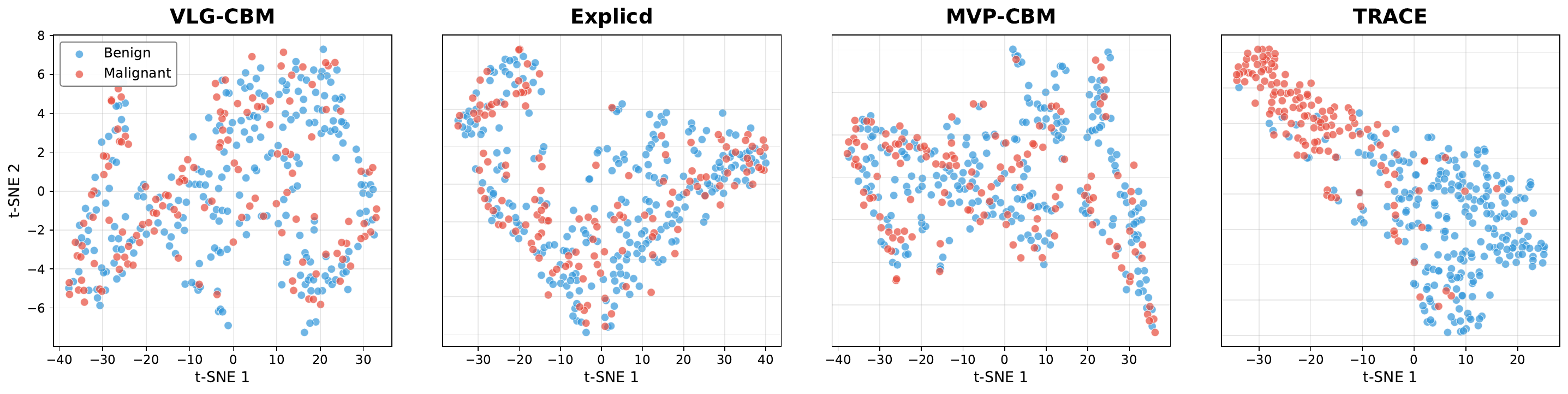}
  \caption{t-SNE visualization of feature embeddings learned by VLG-CBM, Explicd, MVP-CBM, and TRACE.}
  \Description{Four t-SNE plots visualize the feature embeddings learned by VLG-CBM, Explicd, MVP-CBM, and TRACE, respectively. Each plot shows the distribution and separation of samples from the diagnostic classes in the two-dimensional embedding space.}
  \label{fig:tsne}
\end{figure*}

\subsection{Cross-Domain Generalization}

Table~\ref{tab:complete_results} reports zero-shot cross-domain results, where models are trained on BUSC-BUSBRA and directly evaluated on external target domains without fine-tuning. All target datasets, including BUSC-BUSI647$^*$, Ardakani, BUS\_UC, and BrEaST, are tested without structured reports, making this setting a stricter evaluation of transferability under realistic image-only deployment.

Overall, TRACE shows strong cross-domain generalization and achieves the best or near-best performance on multiple target datasets. In particular, TRACE obtains the best Accuracy on Ardakani, the best AUC on BUS\_UC, and the best AUC and Accuracy on BrEaST. It also remains highly competitive on direct transfer from BUSC-BUSBRA to BUSC-BUSI647$^*$, achieving the best Accuracy and one of the strongest AUC results. These results suggest that TRACE does not simply exploit report availability during training, but learns a transferable concept correction mechanism that remains effective without reports at test time. By contrast, pure image models can be competitive on individual targets, while existing concept bottleneck and multimodal methods do not show consistent advantages across domains. This indicates that standard concept modeling or image--text alignment is insufficient for the asymmetric clinical scenario where structured reports are available only during training. With privileged concept supervision, self-editing distillation, and concept-missing augmentation, TRACE achieves more robust generalization across shifts in country, device, and clinical center.

\begin{table}[t]
\centering
\scriptsize
\caption{Ablation on supervision ratio on BUSC-BUSBRA. We vary the proportion of training samples with structured concept supervision from 100\% to 50\% and 20\%. Best and second-best results are highlighted in \textcolor{red}{red} and \textcolor{blue}{blue}.}
\label{tab:ablation_ratio}
\setlength{\tabcolsep}{3.5pt}
\renewcommand{\arraystretch}{1.03}
\resizebox{\columnwidth}{!}{%
\begin{tabular}{llccc}
\toprule
\textbf{Editor} & \textbf{Ratio} & \textbf{AUC} & \textbf{Acc} & \textbf{F1} \\
\midrule
\multirow{3}{*}{\textbf{MLP}}
& 100\% & \second{0.9280$\pm$0.0078} & \best{0.8834$\pm$0.0173} & \best{0.8136$\pm$0.0150} \\
& 50\%  & \best{0.9322$\pm$0.0188}   & \second{0.8754$\pm$0.0126} & \second{0.7971$\pm$0.0318} \\
& 20\%  & 0.9265$\pm$0.0192          & 0.8733$\pm$0.0178          & 0.7966$\pm$0.0206 \\
\midrule
\multirow{3}{*}{\textbf{Attention}}
& 100\% & 0.9263$\pm$0.0118          & \best{0.8807$\pm$0.0120} & \best{0.8070$\pm$0.0211} \\
& 50\%  & \best{0.9304$\pm$0.0132}   & \second{0.8770$\pm$0.0165} & \second{0.8035$\pm$0.0215} \\
& 20\%  & \second{0.9272$\pm$0.0127} & \second{0.8770$\pm$0.0098} & 0.8033$\pm$0.0103 \\
\midrule
\multirow{3}{*}{\textbf{Gating}}
& 100\% & \best{0.9311$\pm$0.0182}   & \best{0.8743$\pm$0.0190} & \best{0.7981$\pm$0.0261} \\
& 50\%  & 0.9260$\pm$0.0139          & \second{0.8722$\pm$0.0230} & \second{0.7869$\pm$0.0307} \\
& 20\%  & \second{0.9301$\pm$0.0123} & 0.8684$\pm$0.0148          & 0.7855$\pm$0.0276 \\
\bottomrule
\end{tabular}%
}
\end{table}

\subsection{Ablation Study}

We further analyze TRACE from three perspectives: editor comparison, supervision settings, and supervision scarcity.

First, Table~\ref{tab:ablation_cross_editor} compares different editors under cross-domain evaluation. TRACE shows stable transferability across editor choices, with TRACE-Att achieving the best or tied-best results on multiple target domains. TRACE-MLP achieves higher AUC on some domains, while TRACE-Gate remains competitive in Accuracy, indicating different transfer preferences among editors. Attention provides the most reliable overall choice.

Second, Table~\ref{tab:ablation_combined} compares editors under Full and Missing settings. All editors perform well, suggesting that TRACE gains mainly come from teacher-guided concept editing rather than a specific architecture. Missing improves several metrics on BUSC-BUSBRA, while Full performs better on BUSC-BUSI647, indicating that missing perturbation is dataset-dependent.

Finally, Table~\ref{tab:ablation_ratio} evaluates reduced supervision from 100\% to 50\% and 20\%. TRACE remains effective with partial concept supervision. Attention and Gating show more stable trends, while MLP achieves a higher upper bound under full supervision.

\subsection{t-SNE Visualization}

To further examine representation quality, Fig.~\ref{fig:tsne} visualizes the feature embeddings of VLG-CBM, Explicd, MVP-CBM, and TRACE using t-SNE, where blue and red points denote benign and malignant samples, respectively. TRACE shows clearer class separation, with more compact intra-class clusters and less overlap between the two classes. In contrast, VLG-CBM, Explicd, and MVP-CBM exhibit more cross-class mixing and scattered feature distributions. This suggests that TRACE learns more discriminative representations for breast ultrasound diagnosis.

This result provides representation-level support for TRACE. The improved separability indicates that teacher-guided concept editing transfers structured report supervision into more robust image-only representations, which is consistent with the superior AUC, Accuracy, and F1 performance in the main results.

\section{Conclusion}

In this paper, 
we presented TRACE, a training-time report-guided framework for robust breast ultrasound diagnosis under incomplete concepts. Rather than treating structured reports as an inference-time modality, TRACE uses them as privileged concept teachers during training to learn how coarse image-derived concepts should be refined. By integrating teacher-guided concept editing, a clinically ordered concept space, Strategic Concept Missing Training, and an image-only self-editor, TRACE enables report-free diagnosis at test time while preserving clinically meaningful concept refinement. We also introduced BUSC, a concept-enriched benchmark that aligns ultrasound images, diagnostic labels, and structured clinical concepts. Extensive experiments in both in-domain and zero-shot cross-domain settings show that TRACE consistently outperforms strong baselines and supports robust image-only deployment across datasets. These findings indicate that learning concept correction from training-only structured supervision is a practical and effective approach to clinically grounded and generalizable breast ultrasound diagnosis.

\label{sec:conclusion}

\begin{acks}
This work is supported by the China Postdoctoral Science Foundation (Grant No. 2025M781597), the Hubei Provincial Natural Science Foundation of China (Grant No. 2026AFB700), the Fundamental Research Funds for the Central Universities, China (Grant No. XJ2026000901), and the Academy of Frontier Interdisciplinary Research at Central China Normal University.
\end{acks}

\bibliographystyle{ACM-Reference-Format}
\bibliography{ref}

% %%
% %% If your work has an appendix, this is the place to put it.
% \appendix

% \section{Additional Experimental Details}
% \label{app:experimental-details}

% \section{Complete Preliminary Results}
% \label{app:preliminary-results}

\end{document}